\documentclass[10pt,twocolumn,letterpaper]{article}%

\usepackage{cvpr}
\usepackage{times}
\usepackage{epsfig}
\usepackage{graphicx}
\usepackage{amsmath}
\usepackage{amssymb}

\usepackage{color}
\definecolor{green}{rgb}{0, 0.5, 0}
\definecolor{orange}{rgb}{0.8, 0.6, 0.2}
\definecolor{red}{rgb}{1.0, 0.0, 0.0}
\definecolor{teal}{rgb}{0.0, 0.4, 0.4}
\definecolor{purple}{rgb}{0.65,0,0.65}
\definecolor{saffron}{rgb}{0.95,0.75,0.2}
\definecolor{turquoise}{rgb}{0.0,0.5,0.5}

\usepackage[ruled,vlined,linesnumbered]{algorithm2e}
\usepackage{multirow}

\newcommand{\kx}[1]{{\color{black}#1}}
\newcommand{\new}[1]{{\color{black}#1}}

\usepackage[normalem]{ulem}

\usepackage{mathtools}
\usepackage{amsmath, amsthm, amssymb, amsfonts, amsopn}
\usepackage{graphicx} 
\usepackage{textcomp}
\usepackage{xfrac}
\usepackage{bbm}

\usepackage{overpic}
\usepackage{subfig}
\usepackage{wrapfig}

\newcommand{\hidecomment}[1]{}

\renewcommand{\vec}[1]{\mathbf{#1}}

\newcommand{\R}{\mathbb{R}}

\usepackage[pagebackref=true,breaklinks=true,letterpaper=true,colorlinks,bookmarks=false]{hyperref}

\cvprfinalcopy 

\def\cvprPaperID{4311} 

\ifcvprfinal\fi

\begin{document}

\title{Shape2Motion: Joint Analysis of Motion Parts and Attributes from 3D Shapes}

\author{
Xiaogang Wang$^{1}$ \quad Bin Zhou$^1$
\quad Yahao Shi$^1$ \quad Xiaowu Chen$^1$ \quad Qinping Zhao$^1$
\quad Kai Xu$^{2}$\thanks{Corresponding author: kevin.kai.xu@gmail.com}\\
$^1$Beihang University \quad\quad $^2$National University of Defense Technology\\
}

\maketitle

\begin{abstract}

For the task of mobility analysis of 3D shapes, we propose joint analysis for simultaneous motion part segmentation and motion attribute estimation, taking a single 3D model as input. The problem is significantly different from those tackled in the existing works which assume the availability of either a pre-existing shape segmentation or multiple 3D models in different motion states.
To that end, we develop Shape2Motion which takes a single 3D point cloud as input,
and jointly computes a mobility-oriented segmentation and the associated motion attributes.
Shape2Motion is comprised of two deep neural networks designed for mobility proposal generation and mobility optimization, respectively. The key contribution of these networks is the novel motion-driven features and losses used in both motion part segmentation and motion attribute estimation. This is based on the observation that the movement of a functional part preserves the shape structure. We evaluate Shape2Motion with a newly proposed benchmark for mobility analysis of 3D shapes. Results demonstrate that our method achieves the state-of-the-art performance both in terms of motion part segmentation and motion attribute estimation.

\end{abstract}

\section{Introduction}
The analysis of part mobilities is a key step towards
function analysis of 3D shapes~\cite{HU:2018:func_star},
finding numerous potential applications in robot-environment interaction~\cite{myers2015affordance,nguyen2016detecting}.
In this work, we approach this problem from a data-driven perspective:
Given a 3D shape as static observation, learn to simultaneously segment the shape
into motion parts and estimate associated motion attributes (type and parameters).

\begin{figure}[t!]\centering
  \begin{overpic}[width=1.0\linewidth,tics=10]{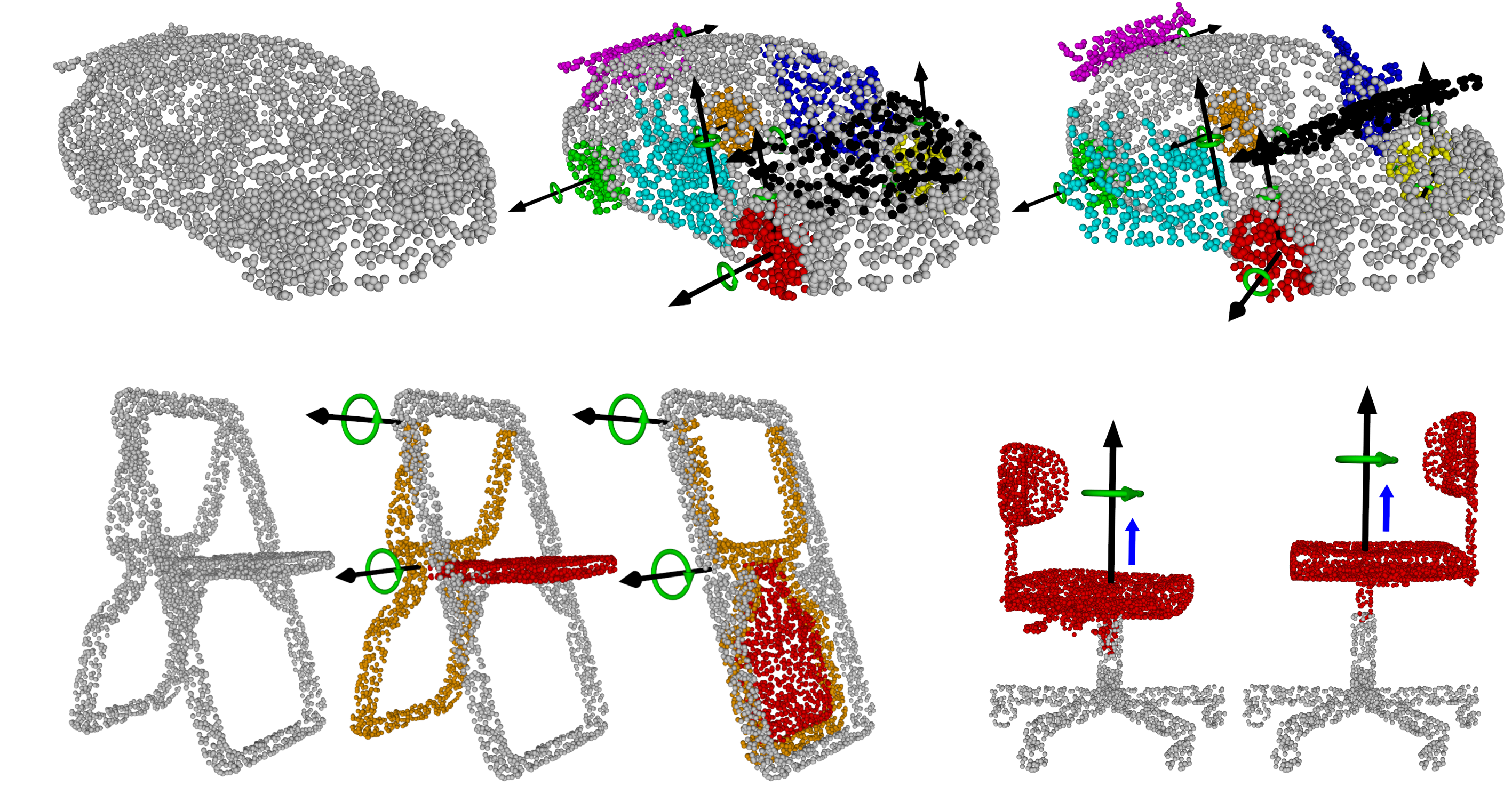}
  \put(50,29){\small (a)}
  \put(25,-2){\small (b)}
  \put(78,-2){\small (c)}
  \end{overpic}
  \caption{Three examples of simultaneous motion part segmentation and motion attribute inference. In each example, the left is the input point cloud; the middle demonstrates the analyzed result (segmented motion parts are shaded in distinct colors and their associated motions depicted with arrows); the right shows the motion parts are moved according to the analyzed motions.}
  \label{fig:teaser}\vspace{-8pt}
\end{figure}

In most existing approaches, mobility analysis is conducted on the pre-segmented parts of a 3D model~\cite{hu2017learning}.
However, the pre-existing segmentation of a 3D model does not necessarily conform to its mobilities.
For example, in many car models, the door is not a separate component (Figure~\ref{fig:teaser}(a)).
This greatly limits the utility of those methods
on existing shape repositories where mobility-oriented segmentation is often unavailable.
Another type of method extracts mobility through comparing multiple different motion states of a single object,
e.g., a scissor with different opening angles~\cite{yi2018deep}.
Such requirement is, unfortunately, hard to meet for most testing shapes.
Moreover, not all mobility can be easily spot by motion state comparison.
For instance, it would be very hard to capture a rotating sphere without very accurate slippage analysis~\cite{gelfand2004shape}.
Last but not least, existing methods can only output a single mobility for each motion part, while
in reality one part may possess multiple motions (e.g., the seat of a swivel chair in Figure~\ref{fig:teaser}(c)).

\begin{figure}[t]
  \centering
  \includegraphics[width=0.85\linewidth]{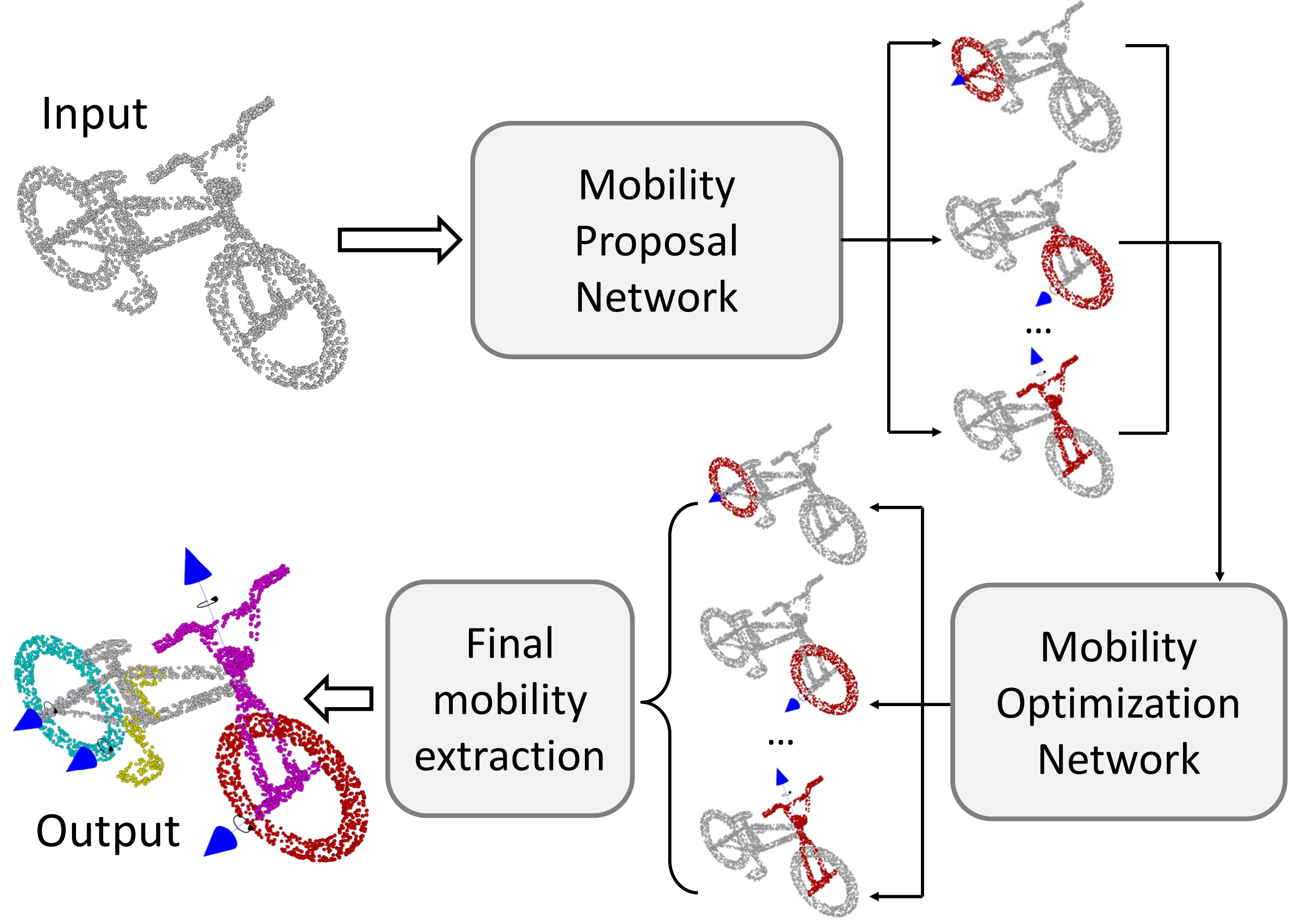}
  \caption{An overview of the three stages of our method.}
  \label{fig:over_view_all}
  \vspace{-8pt}
\end{figure}

We propose Shape2Motion, a method that consumes a single 3D shape in point cloud as input,
and jointly computes a mobility-oriented segmentation and estimates the corresponding motion attributes.
Shape2Motion adopts a propose-and-optimize strategy, in a similar spirit to the proposal-based object detection from images~\cite{ren2015faster}.
It consists of two carefully designed deep neural networks, i.e., a mobility
proposal network (MPN) followed by a mobility optimization network (MON).
MPN generates a collection of mobility proposals and selects a few high-quality ones. Each mobility proposal is comprised of a motion part and its associated motion attributes; the latter refers to motion type (\emph{translation}, \emph{rotation} and \emph{translation+rotation}) and motion parameters (\emph{translation direction} and \emph{rotation axis}).
MON optimizes the proposed mobilities through jointly optimizing the motion part segmentation and motion attributes.
The optimized mobilities are then merged, yielding the final set of mobilities.
Figure~\ref{fig:over_view_all} shows an overview of our method.

Our key insight in designing these networks is to fully exploit the coupling between a motion part
and its mobility function: The movement of a functional part does not break the shape structure.
See the examples in Figure~\ref{fig:teaser}: the opening or closing of a car door keeps it hinged on the car frame; the folding or unfolding of a folding chair preserves the attachment relations between its constituent parts.
Therefore, we propose a motion-driven approach to encode mobilities and to measure the correctness (or loss)
of motion part segmentation and motion attribute inference.
Given a mobility proposal, we first move the part according to
the corresponding motion and then inspect how much the movement preserves the shape structure.
Given the ground-truth part mobility, we can measure
the pose deviation of the moved part from that of its ground-truth counterpart under ground-truth motion.
Motion-driven pose loss amplifies the direct loss of segmentation and motion parameters, thus greatly
improves the training convergence.

We perform extensive evaluations of Shape2Motion over a newly proposed benchmark for mobility analysis of 3D shapes. Results demonstrate that our method achieves the state-of-the-art performance both in terms of motion part segmentation and motion attribute estimation, and show the advantages of our design choices over several baselines.
Our work makes the following contributions:
\begin{itemize}
\vspace{-4pt}
  \item We propose the problem of joint analysis for motion part segmentation and motion attribute prediction from a single 3D shape.
\vspace{-6pt}
  \item We design the first deep learning architecture to approach the above problem with two carefully designed networks, responsible for mobility proposing and mobility optimization, respectively.
\vspace{-6pt}
  \item We contribute the first benchmark of 3D shape mobility analysis, encompassing both motion part segmentation and motion attribute estimation.
\end{itemize}

\section{Related Work}

\begin{figure*}[t]
  \centering
  \includegraphics[width=\textwidth]{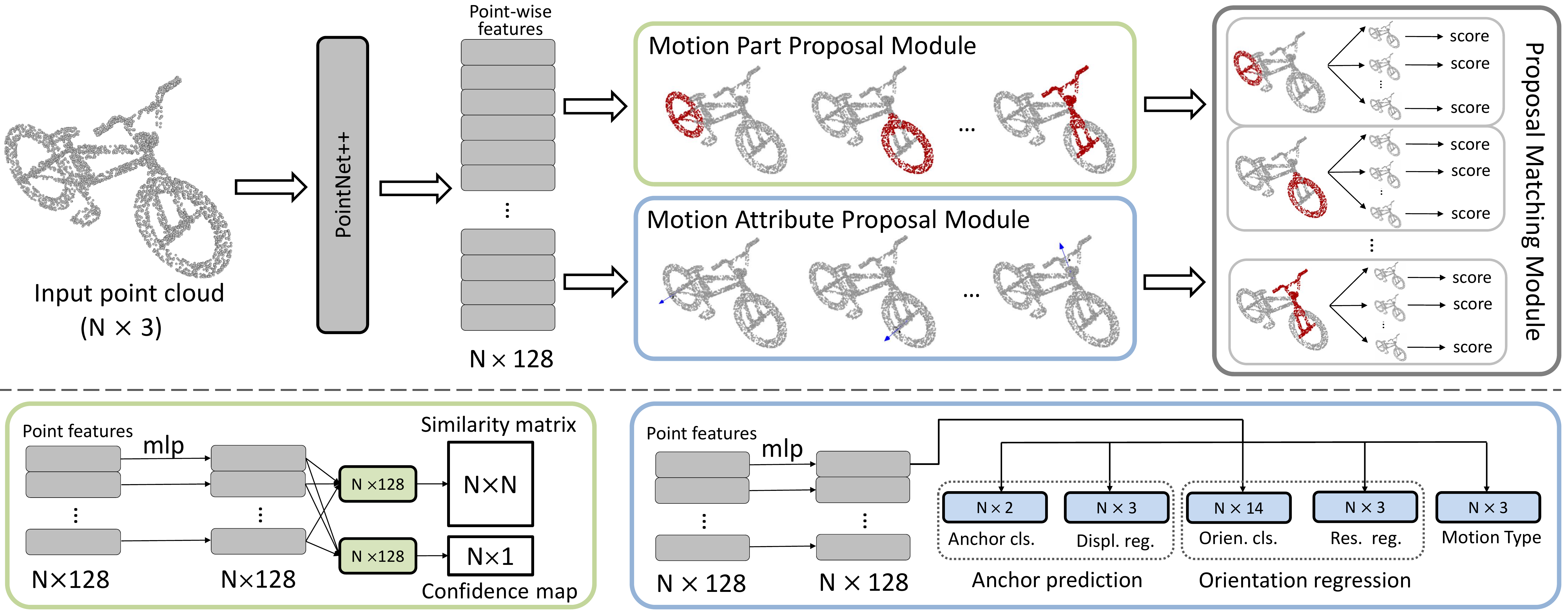}
  \caption{Architecture of Motion Proposal Network (MPN). MPN contains three modules:
  a Motion Part Proposal Module (green boxes), a Motion Attribute Proposal Module (blue boxes) and a Proposal Matching Module (grey box to the top-right).}
  \label{fig:Overview_net}
  \vspace{-8pt}
\end{figure*}

\paragraph{Mobility analysis of 3D shapes.}
Dynamic functionalities of articulated 3D objects can be characterized by the motion of one or more of their constituent parts, which is commonly referred to as part mobility~\cite{HU:2018:func_star}.
There have been a few works on discovering part mobility from an input object~\cite{xu2009joint,mitra2010illustrating,sharf2014mobility},
or from a sequence of RGBD scans of the dynamic motion of an articulated model~\cite{li2016mobility}.
Hu et al.~\cite{hu2017learning} propose a method for inferring part mobilities
of a static 3D object by learning from motion sequences of parts from different classes of objects.
Common to these existing approaches is the assumption of a pre-existing part segmentation, which, however, does not necessarily conform to part mobilities. Our method learns to simultaneously segment a 3D shape into motion parts and infer their motion attributes. Moreover, our method does not need a motion sequence of either object or parts, during either training or testing.

\paragraph{Semantics- / function- / mobility-induced segmentation of 3D shapes.}
Semantic segmentation of 3D shapes has gained significant research progress in recent year, benefiting from the advanced machine learning techniques~\cite{Kalogerakis_SG10,sidi2011unsupervised,Wang_SA13,Xie_SGP14} (see also the comprehensive survey in~\cite{xu2016data}), and especially from the powerful feature learning of deep-learning models~\cite{qi2017pnpp,yi2017syncspeccnn,wang2017cnn,kalogerakis20173d}.
The goal of semantic segmentation is to decompose a 3D shape into parts which are meaningful from the assembling or functional point of view. A semantically meaningful part does not necessarily imply a mobility.

Inferring functional parts is another promising way of 3D shape segmentation.
Pechuk et al.~\cite{pechuk2008learning} introduce a supervised method to recognize the functional regions of a shape according to a model of functionality of the shape class.
Kim and Sukhatme~\cite{kim2014semantic} propose to learn a classifier of regions of shapes with functional labels such as ``graspable'', ``liftable'', and ``pushable'' in a supervised manner.
Similarly, Laga et al.~\cite{laga2013geometry} introduce a supervised method for labeling shape parts that with functional tags, based on both geometric features and the context of parts.
The method of Hu et al.~\cite{hu2016learning} defines weight fields over a point-sampled surface, based on a learned classifier predicting the probability of a point belonging to a specific functional region, which could be used to infer a function-induced segmentation.

Mobility-induce segmentation is relatively less studied. The slippage analysis approach of Gelfand and Guibas~\cite{gelfand2004shape} segments a triangle mesh into kinematic surfaces, which indicate regions of the shape that undergo a similar type of motion. Each resulting segment is classified into one of a few types of kinematic surfaces, such as planes, spheres, and cylinders, which are not yet parts with functional mobilities. There is a large body of literature on co-segmentation of a sequence of animated meshes (e.g.~\cite{lee2006segmenting,rosman2012group,ghosh2012deformations}). These works typically handle organic objects (e.g., human and animal) undergoing smooth surface deformation.
In contrast, Yi et al.~\cite{yi2018deep} studies mobility analysis of man-made objects through comparing multiple different motion states of an object, e.g., a scissor with different opening angles.
Our work infers part mobilities from a single, static 3D model, through joinly learning motion part segmentation and motion attribute regression from a large annotated dataset.

\section{Mobility Proposal Network (MPN)}
\label{sec:method}
\paragraph{Terminology and notations.}
Given a 3D shape represented as point cloud $P=\{p_i\}_{i=1}^N$ ($p_i \in \R^3$), our goal is to extract a set of \emph{part mobilities}: $\mathcal{M}=\{M_k\}_{k=1}^K$. Each part mobility is a tuple: $M_k=\langle P_k, A_k \rangle$, which consists of a \emph{motion part segmentation}, $P_k \subset P$, and the corresponding \emph{motion attributes}, $A_k = \langle t_k, m_k \rangle$, where $t_k$ is the \emph{motion type} and $m_k$ the \emph{motion parameter} of the motion.
We consider three motion types: translation (T), rotation (R), and rotation+translation (RT).
Motion parameter is a line in $\R^3$ along the motion axis referring to either a translation direction or a rotation axis, or both. Other motion parameters, such as motion range or joint limit, are left for future works.

Our Mobility Proposal Network (MPN) is designed to generate a collection of quality proposals of part mobilities, $\mathcal{M}^\text{P}_k=\{M^\text{P}_k\}$, so that the final part mobilities could be selected from them via an optimization stage in Section~\ref{sec:mon}.
MPN itself is composed of three modules: a \emph{motion part proposal module}, a \emph{motion attribute proposal module} and a \emph{proposal matching module}.
The first two modules are devised to propose motion part segmentations and the corresponding motion attributes, respectively.
The third module is used to select a set of good-quality mobility proposals, through matching the two kinds of proposals based on a motion-driven filtering scheme.

\subsection{Motion Part Proposal Module}
\label{subsec:mppm}
Unlike the existing approaches, we do not assume the availability of mobility-induced shape segmentation, but instead try to solve it through coupling it with the inference of motion attributes.
First of all, off-the-shelf shape segmentation methods cannot be used since a semantically meaningful part may not correspond to a motion part.
Meanwhile, it is quite challenging to cast motion part segmentation into a plain labeling problem, as what is done in semantic segmentation, since the possible categories of motion parts could be extremely large, considering the combination of different functions and various motions.


We therefore opt for a proposal-based solution and devise a motion part proposal network based on the recently proposed SGPN (Similarity Group Proposal Network)~\cite{SGPN_17}.
SGPN is developed for object proposal in scene segmentation.
It regresses a point similarity matrix $S$ to accentuate point group proposals, from which object instances could be extracted. Each entry $S_{ij}$ in the similarity matrix indicates whether two points, $p_i$ and $p_j$, belong to the same object instance. Each row can then be viewed as an object proposal. They also regress a confidence score for each object proposal. In what follows, we explain the main adaptions we make for our problem setting.

\paragraph{Similarity matrix.}
To achieve motion part proposing, we define a motion-based point similarity matrix $S^\text{M} \in \R^{N\times N}$ to encode whether two points belong to the same motion part. We train a fully-connected network to estimate $S^\text{M}$, based on the motion similarity matrix loss:
\begin{equation}
  L_\text{sim}=\sum_{p_i,p_j\in P, i\neq j}l_{i,j},
\label{eq:simloss}
\end{equation}
where
$$ l_{i,j}=\left\{
\begin{array}{rcl}
\|F(p_i)-F(p_j)\|_2,     &      {mp(i,j)=1}\\
\max\{0,K-\|F(p_i)-F(p_j)\|_2\},      & {mp(i,j)=0}
\end{array} \right. $$
where $mp(i,j)$ indicates whether $p_i$ and $p_j$ belong to the same motion part in the ground-truth ($1$ means `yes').
$F$ is point-wise feature computed with PointNet++~\cite{qi2017pnpp}.
$K$ is a constant that controls the degree of motion-based dissimilarity in a proposal; we use $K=100$ by default.
The rows of $S^\text{M}$ then represent $N$ different motion part proposals.
During testing, we use a threshold $\tau_\text{sim}$ to binarize the regressed similarity matrix so that each row of the matrix represents a binary segmentation of motion part.
We set $\tau_\text{sim}=100$ throughout our experiments.

\paragraph{Confidence map.}
To rate the quality of the motion part proposals in $S^M$,
we regress a confidence map $C \in \R^{N\times1}$ where $i$-th entry corresponds to $i$-th motion part proposal in $S^\text{M}$.
The ground-truth confidence values in $C^\text{gt}$ are computed as the IoU between the point set of a proposed motion part $P_i$ and its ground-truth counterpart $P^\text{gt}_i$.
The loss for training the fully-connected regression network, $L_\text{conf}$, is the mean-square-error (MSE) loss between $C$ and $C^\text{gt}$.
During testing, only those proposals with a confidence value higher than $\tau_\text{conf}=0.5$ are considered as valid ones.

\subsection{Motion Attribute Proposal Module}
\label{subsec:mapm}
Existing works usually estimate a single motion for each motion part.
In reality, however, a motion part can possess multiple different motions.
Taking the models in Figure~\ref{fig:teaser} for example, the front wheel of a car can rotate along two different axes (steering and rolling); the seat of a swivel chair goes up and down (translating) while rotating.

\begin{figure}[t]
  \centering
  \begin{overpic}[width=\linewidth,tics=10]{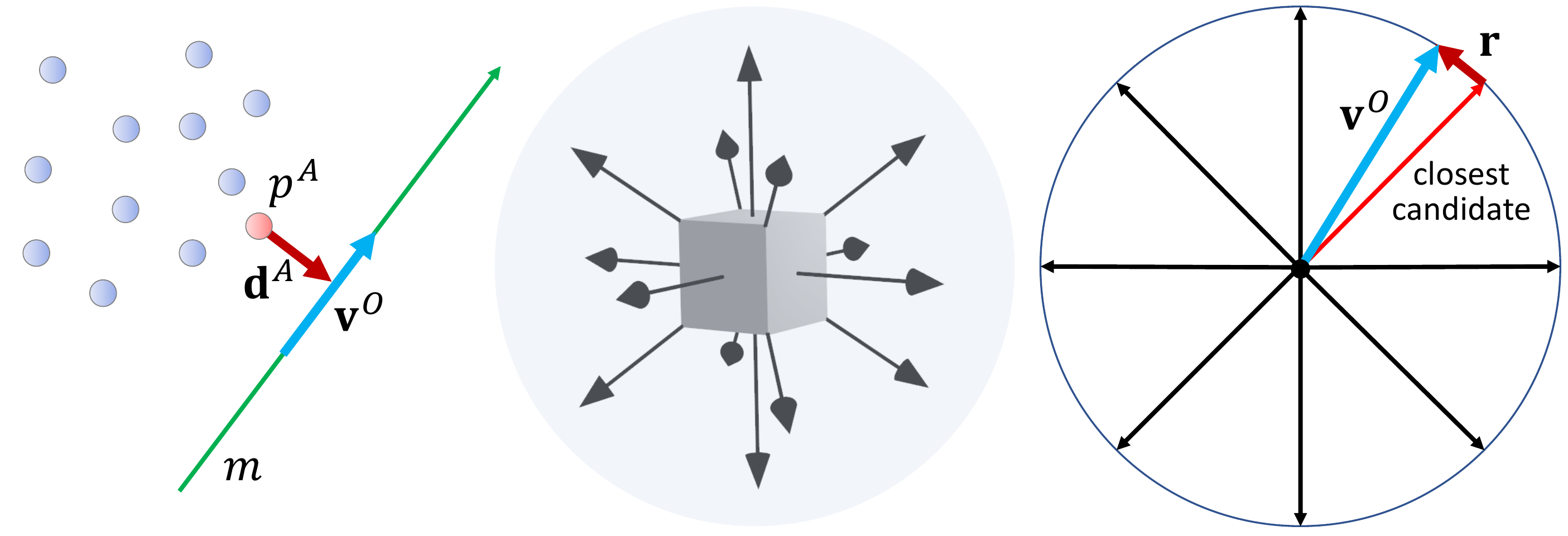}
  \put(12,-3){\small (a)}
  \put(46,-3){\small (b)}
  \put(81,-3){\small (c)}
  \end{overpic}
  \caption{Illustrations of anchor-based motion axis parameterization (a), orientation discretization (b),
  and orientation prediction via classification and residual regression (c).}
  \label{fig:Regress_vectors}
  \vspace{-8pt}
\end{figure}

Instead of estimating several motions for each motion part proposal, which can cause combinatorial explosion,
we opt for an independent motion attribute proposal, based on a motion attribute proposal module.
Given a 3D point cloud $P$, this module proposes a set of motion attributes $\{\langle t, m \rangle\}$. Each attribute contains a \emph{motion type} $t$ and a motion axis line $m$. The latter is parameterized by an \emph{anchor point} $p^\text{A}\in P$, a \emph{displacement vector} $\vec{d}^\text{A}$, and an \emph{orientation vector} $\vec{v}^\text{O}$.
By selecting a point in the point cloud as anchor, the motion axis line $m$ can be uniquely determined by a displacement to the anchor $\vec{d}^\text{A}$ and an orientation $\vec{v}^\text{O}$ (Figure~\ref{fig:Regress_vectors}(a)).

Figure~\ref{fig:Overview_net} summarizes the network architecture of motion attribute proposal module.
It contains five branches: two for anchor prediction, two for orientation regression and one for motion type prediction. Below we explain them in detail.

\paragraph{Anchor prediction.}
To spatially pin down a motion axis, we train a network to select a point in the input point cloud that is closest to the line along a ground-truth motion axis.
\kx{We use a binary indicator vector to encode point selection.}
In addition, we regress the displacement vector between the anchor point and the ground-truth line; see Figure~\ref{fig:Regress_vectors}(a).
By doing this, motion axes prediction is invariant to the shape pose.
The anchoring loss is computed as:
\begin{equation}
   L_\text{anchor}=L_\text{ap}+L_\text{dis},
\label{eq:multi_task_loss}
\end{equation}
where $L_\text{ap}$ is softmax loss for binary classification of whether a point is anchor or not, and $L_\text{dis}$ an L2 loss between the predicted displacement vector and ground-truth.

\begin{figure*}[ht]
  \centering
  \includegraphics[width=\textwidth]{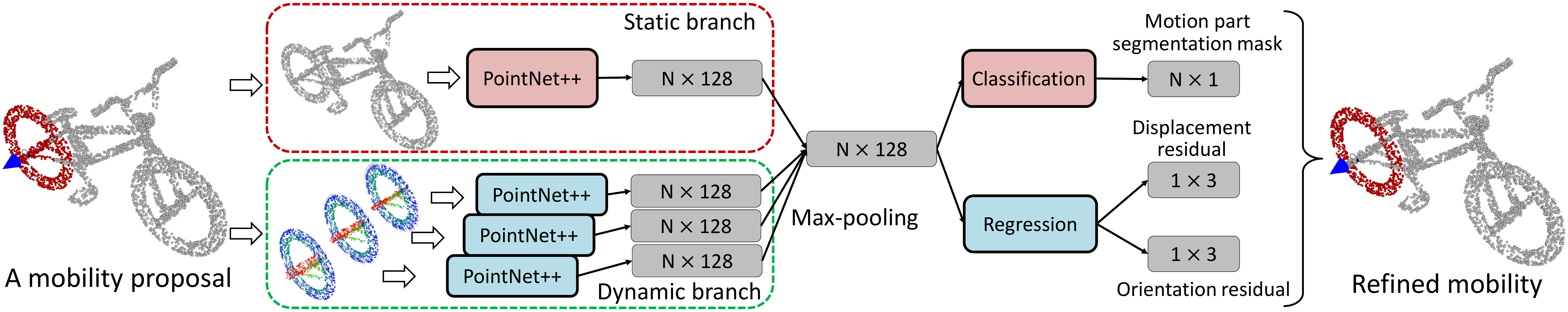}
  \caption{Architecture of Motion Optimization Network (MON).
  Given a part mobility, MON encodes both static shape geometry (red box) and dynamic part motion (green box). It produces a refined segmentation of the motion part \kx{(removing the points of ``back fork'' which are mis-labeled as ``wheel'' in the input)} and residual vectors for correcting the motion axis.
}
  \label{fig:Overview_optimization_net}
  \vspace{-8pt}
\end{figure*}

\paragraph{Orientation regression.}
Direct regression of orientation is quite difficult to train.
We instead turn the problem into a classification problem through discretizing the space of orientation into $14$ candidates.
For each candidate orientation, the network estimates both a classification probability and a residual vector used to correct the error caused by the discretization; see Figure~\ref{fig:Regress_vectors}(b) for illustration.
The loss of orientation regression is:
\begin{equation}
   L_\text{orien}=L_\text{class}+L_\text{res},
\label{eq:multi_task_loss}
\end{equation}
where $L_\text{class}$ is the softmax classification loss and
$L_\text{res}$ an L2 loss between the estimated residual vector and ground-truth.

\paragraph{Motion type prediction.}
Another network is trained to classify the mobility into one of the three motion types, with the loss
$L_\text{type}$ being a softmax classification loss.

\subsection{Proposal Matching Module}
\label{subsec:mpsm}
Having obtained a set of motion part proposals and a set of motion attribute proposals,
this module selects the best combinations of the two, leading to a set of high-quality mobility proposals.
To do so, for each motion part proposal, we find a set of motion attribute proposals that best matches it.
This is achieved by training another network, which takes the feature maps of a motion part and a motion attribute learned from the respective modules, and regresses a matching score for them.
To train this network, we compute a matching score loss between a proposed mobility $M^\text{prop}$ and its ground-truth counterpart $M^\text{gt}$: $L_\text{ms}=|S^\text{pred}-S^\text{gt}|$. $S^\text{pred}$ is the matching score predicted by the network. $S^\text{gt}=score(M^\text{prop}, M^\text{gt})$, where $score$ is a scoring function measuring the similarity between a proposed mobility and the ground-truth.

To measure the similarity between two mobilities, we propose a motion-driven metric which accounts for both the similarity of motion part segmentation and motion attributes. The basic idea is to let the motion parts move a prescribed amount according to their respective motions, for both mobilities, and measure the degree of spatial alignment between the moved parts:
\begin{equation}\label{eq:match_score}\small
  score(M_j, M_k)=\frac{1}{N}\sum_{p_i \in P}\|move(p_i,M_j)-move(p_i,M_k)\|_2,
\end{equation}
where $move(p_i, M_k)$ is a moving function:
$$
move(p_i,M_k)=\left\{
\begin{array}{rcl}
T(p_i, m_k, \delta)/dist(p_i, m_k), & {p_i \in P_k}\\
0, & {p_i \notin P_k}
\end{array}\right.
$$
which moves point $p_i \in P$ according to the motion defined in $M_k$, if $p_i$ belongs to the point cloud of the corresponding motion part.
$T(p_i, m_k, \delta)$ transforms a point according to the motion axis line by an amount of $\delta$.
$\delta$ takes $15\%$ of the diagonal length of $P$'s bounding box for translation, and $90^\circ$ for rotation.
$dist(p_i, m_k)$ is the distance from $p_i$ to motion axis $m_k$, which is used to normalize the movement.

\section{Mobility Optimization Network}
\label{sec:mon}

Since the proposals for motion parts and attributes are generated separately, they could be inaccurate and insufficiently conforming with each other. Mobility Optimization Network (MON) is trained to jointly optimize both \kx{in an end-to-end fashion} (Figure~\ref{fig:Overview_optimization_net}).
Given a proposed mobility, MON refines it through predicting a binary segmentation of motion part out of the input shape point cloud, and regressing two residual vectors for correcting the displacement $\vec{d}^\text{A}$ (w.r.t. the anchor point) and orientation $\vec{v}^\text{O}$ of the motion axis respectively.

To account for both shape geometry and part motion, MON takes both static and dynamic information as input. The static branch encodes the shape point cloud with PointNet++ features.
The dynamic branch takes as input a bunch of moved point clouds corresponding to the motion part being refined.
Specifically, we move the part about the motion axis by three specific amounts ($5\%$, $10\%$ and $15\%$ of the diagonal length of shape bounding box for translation and $30^\circ$, $60^\circ$ and $90^\circ$ for rotation). The moved point clouds are again encoded with PointNet++ except that the point-wise moving vectors are also encoded (using the point normal channel).
\kx{Note that the two branches do not share weights.}
By integrating both static geometric features and \emph{motion-driven} dynamic features, MON achieves highly accurate segmentation and regression.

The loss for MON is composed of two parts, a point-wise labeling loss and the residual losses for motion axis:
\begin{equation}
\begin{split}
   L_\text{MON}(P,M_k) &= \sum_{p_i \in P}{L_\text{label}(p_i)}\\
                &+ L_\text{res}(r_\text{displ}^\text{pred},r_\text{displ}^\text{gt}) + L_\text{res}(r_\text{orien}^\text{pred},r_\text{orien}^\text{gt}),
\end{split}
\label{eq:mon_loss}
\end{equation}
where $L_\text{label}$ is a negative log-likelihood loss for labeling and $L_\text{res}$ takes L2 loss.
The ground-truth residuals are computed by comparing the predicted displacement/orienation vectors and their ground-truth counterparts.

\section{Final Mobility Extraction}
\label{sec:final}
During the testing phase, having obtained a set of high-quality mobility proposals with associated matching scores, we also need a final extraction process to merge the proposals as output.
This process is similar to the Non-Maximum Suppression (NMS) step employed in many proposal-based object detection works.
Different from NMS, however, we need to select both the motion part and its associated motion attributes. Furthermore, one part may have multiple possible motions.

We first select motion parts: When multiple motion part proposals overlap significantly, we keep only the one with the highest matching score. Then for each selected motion part, we select distinct motion attributes from those mobility proposals whose motion part has a significantly large IoU against that part.
For translation, only one direction with the highest score is selected, assuming a part can slide along only one direction.
For rotation, we perform a greedy selection of high score axes while ensuring the angle between every two selected axes is larger than a threshold ($45^\circ$).

\section{Details, Results and Evaluations}
\label{sec:opti}

\subsection{Network training}
\label{subsec:training}

\paragraph{Training scheduling.}
Our network is implemented with Tensorflow.
We used Adam~\cite{Kingma_2014} for training and set the initial learning rate to $0.001$.
Our network is trained in three stages.
In the \emph{first stage}, we train the Motion Part Proposal Module and Motion Attribute Proposal Module, each for $100$ epochs. The mini-batch size is $8$ shapes.
In the \emph{second stage}, we train the Proposal Matching Module using the motion part proposals and motion attribute proposals generated from the first stage, again, for $100$ epochs.
Here, we use mini-batches of size $64$ ($4$ shapes $\times 16$ motion part proposals).
The training motion part proposals are selected from those whose IoU against ground-truth is greater than $0.5$ and within top $25\%$.
The \emph{third stage} trains the Motion Optimization Network for $100$ epoch.
The training mobility proposals are selected from those whose matching error at most $0.05$ and within top $10\%$.
The mini-batch is $8$ mobility proposals.
Our networks make extensive use of PointNet++~\cite{qi2017pnpp}, for which we use the default architecture and parameters provided in the original paper.
\kx{The training and testing time is reported in the supplemental material.}

\paragraph{Data augmentation.}
Given a training dataset with ground-truth mobilities, we perform data augmentation via generating two kinds of shape variations. We first generate a number of \emph{geometric variations} for each training shape based on the method described in~\cite{fu_cgf}. Furthermore, based on the ground-truth mobility in the training shapes, we move the motion parts according to their corresponding motion attributes, resulting in a large number of \emph{motion variations}.
\kx{See the supplemental material for a few samples of our training data augmentation.}

\subsection{Shape2Motion benchmark}
\label{sec:benchmark}

We contribute the first benchmark for mobility analysis, called Shape2Motion, encompassing both motion part segmentation and motion attribute estimation.
It contains $2440$ models in $45$ shape categories.
These shapes are selected from ShapeNet~\cite{ShapeNet2015} and 3D Warehouse~\cite{Tri3Dwarehouse}.
\kx{A overview and detailed information of the benchmark dataset can be found int the supplemental material.}
For each shape, we manually label its motion parts and the corresponding motion attributes, using an easy-to-use annotation tool developed by ourselves.
The benchmark along with the annotation tool will be released (project page:
\url{www.kevinkaixu.net/projects/shape2motion.html}).

\paragraph{Annotation tool.}
Mobility annotation of 3D shapes is tedious. To ease the process, we developed an annotation tool
with an easy-to-use user interface.
The tool consists of two parts, one for motion part annotation and one for motion attribute annotation.
A notable feature of our tool is that it allows the user to visually verify the correctness of a annotated mobility, by animating the annotated motion part with the corresponding motion attributes prescribed by the user.
Using our tool, it takes $80$ seconds in average to annotate a 3D shape.
\kx{In the supplemental material, we provide more details about the annotation tool.}

\begin{figure*}[t!]
  \centering
  \includegraphics[width=0.96\textwidth]{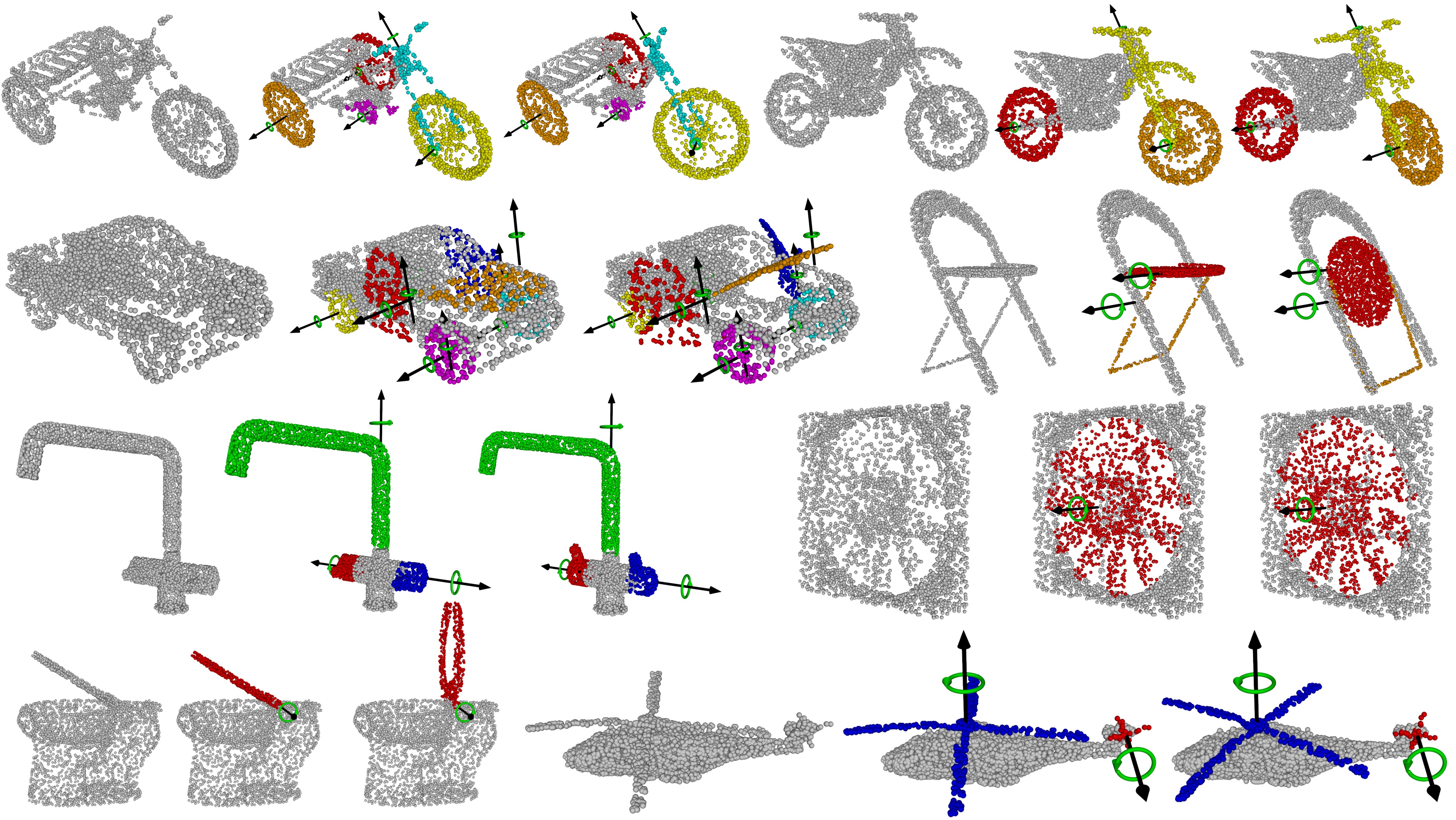}
  \caption{Visual results of mobility analysis. For each example, we show from left to right the input point cloud, the extracted mobilities (motion parts and motion axes) and the point cloud with motion parts moved according to detected mobility.}
  \label{fig:MoreResults_1}
  \vspace{-5pt}
\end{figure*}

\subsection{Evaluation on Shape2Motion}
\label{sec:evals2m}
We train and test our model on the Shape2Motion dataset, with a training / testing split of $8:2$. After data enhancement, we obtain $11766$ shapes with $32298$ mobility parts in total as training data.
Figure~\ref{fig:MoreResults_1} shows a couple of visual results of mobility analysis.
\kx{More results can be found in the supplemental material.}

To compare with the state-of-the-art method in~\cite{yi2018deep}, we train their model on the Shape2Motion dataset.
This model requires shapes in two different motion states as input both for training and testing.
To this end, we compute for each shape another motion state according to the ground-truth mobilities.
Their method is trained to predict motion parts and associated 3D motion flows,
while our method predicts motion parts and motion attributes.
To make the two methods comparable, we convert the mobilities predicted by our method into 3D motion flows.
We measure IoU to evaluate motion part segmentation and End-Point-Error (EPE)~\cite{YanX16} to evaluate 3D motion flow.
The comparison is shown in Table~\ref{tab:shape2motion}.
Our method outperforms theirs because our method can detect those mobilities which are difficult to capture by motion state comparison. For instance, it would be very hard to capture a rotating wheel with two motion states.

\begin{table}[t]\centering
\setlength{\tabcolsep}{1.8mm}{
  \begin{tabular}{l|c|c|c|c|c}
    \hline
     & IoU & EPE & MD & OE  & TA \\
    \hline\hline
    Yi et al.~\cite{yi2018deep}   &61.0 & 0.051 & - & - & -\\
    \hline
    SGPN~\cite{SGPN_17}+BL    &79.4 & - & 0.11 & 3.13 & 0.96\\
    \hline
    Ours (w/o MON)            & 82.3 & 0.028 & \textbf{0.01} & 0.12 & \textbf{0.98}\\
    Ours                       & \textbf{84.7} & \textbf{0.025} & \textbf{0.01} & \textbf{0.11} & \textbf{0.98}\\
    \hline
  \end{tabular}}
  \caption{Comparison on the Shape2Motion benchmark.}
\label{tab:shape2motion}\vspace{-5pt}
\end{table}

In addition, we develop a baseline method (Table~\ref{tab:shape2motion}, row 2) for directly segmenting motion parts and regressing motion attributes.
Specifically, we use the instance segmentation network SGPN~\cite{SGPN_17} to perform motion part segmentation. For motion attribute prediction, we design a baseline network that takes the point cloud of a motion part as input and regress the motion type and motion axis (more details in the supplemental material).
To facilitate comparison, we define three metrics for evaluating motion attributes.
Minimum Distance (MD) and Orientation Error (OE) measure the distance and angle between the predicted motion axis line and the ground-truth. Type Accuracy (TA) is accuracy of motion type prediction.
Our method achieves better performance for both motion part segmentation and motion attribute estimation, through leveraging the synergy between the two tasks in a joint optimization framework.

\begin{table}[t]\centering
\setlength{\tabcolsep}{1.8mm}{
  \begin{tabular}{l|c|c|c|c|c}
    \hline
     & IoU & EPE & MD & OE  & TA \\
    \hline\hline
    Hu et al.~\cite{hu2017learning}   & - & - & 0.030 & 8.14 & -\\
    \hline
    SGPN~\cite{SGPN_17}+BL    & 54.3 & - & 0.145 & 11.8 & 0.74\\
    \hline
    Ours (w/o MON)            & 58.6           & 0.074          & 0.028         & 0.78          & \textbf{0.93}\\
    Ours                      & \textbf{64.7} & \textbf{0.061} & \textbf{0.024} & \textbf{0.12} & \textbf{0.93}\\
    \hline
  \end{tabular}}
\caption{Comparison on the dataset of Hu et al.~\cite{hu2017learning}.}
\label{tab:hudataset}\vspace{-5pt}
\end{table}

\subsection{Evaluation on the dataset in~\cite{hu2017learning} }
\label{subsec:hudata}
We also evaluate our method on the dataset in the work of Hu et al.~\cite{hu2017learning}, which contains $315$ shapes in $37$ categories, with $368$ mobility parts in total.
We again performed data enhancement on this dataset.
Table~\ref{tab:hudataset} reports the comparison of the afore-mentioned metrics
between the method of Hu et al.~\cite{hu2017learning}, the SGPN+baseline method, and our method.

In the method of Hu et al., the availability of segmented motion parts is assumed.
To predict the motion attributes (motion type and motion parameters) of a part, it takes the part and a static part adjacent to that part as input, and perform prediction via metric learning.
We report their performance on two metrics, i.e., MD and OE.
The results shows our method is advantageous thanks to our powerful deep-learning model trained effectively with the motion-driven strategy.
Since their motion types are different from ours, we do not report TA comparison.

\subsection{Analysis of parameters and networks}
\label{subsec:param}

\begin{figure}[t]
  \centering
  \includegraphics[width=0.9\linewidth]{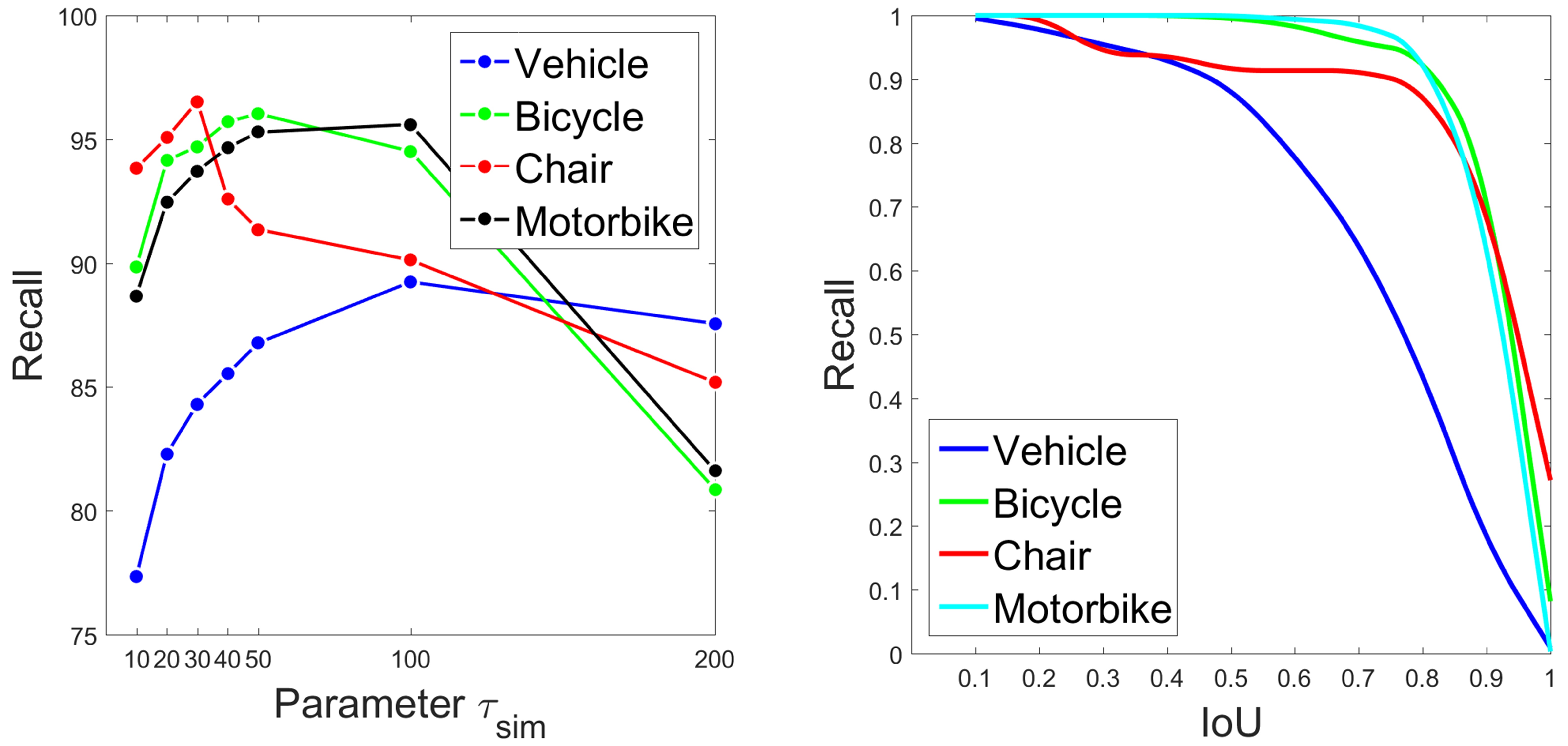}
  \caption{Left: Recall rate of motion parts over increasing $\tau_\text{sim}$, when IoU against ground-truth is fixed to $0.5$. Right: Recall rate over IoU.}
  \label{fig:Recall_over_threshold}\vspace{-5pt}
\end{figure}

\paragraph{Effect of the parameter $\tau_\text{sim}$ and $\tau_\text{conf}$.}
As mentioned in Section~\ref{subsec:mppm}, $\tau_\text{sim}$ is used to binarize the similarity matrix to form a binary segmentation of motion part in each row of the matrix. We use $\tau_\text{sim}=100$ by default.
In Figure~\ref{fig:Recall_over_threshold} (left), we study the quality of motion part proposals over the different values of $\tau_\text{sim}$, over four relatively difficult shape categories.
For a fixed IoU threshold ($0.5$), we find that the recall rate grows fast with the increasing $\tau_\text{sim}$,
and then drops as $\tau_\text{sim}$ continues to increase.
The peak is reached around $\tau_\text{sim}=100$.
This parameter $\tau_\text{sim}$ is strongly correlated with the parameter $K$ in Equation (\ref{eq:simloss}), which is the margin used in defining the hinge similarity loss.


In Section~\ref{subsec:mppm}, $\tau_\text{conf}$ is used to filter motion part proposals based on the confidence map.
In Table~\ref{tab:Performance_all}, we also study the motion part quality (average IoU) over different confidence threshold $\tau_\text{conf}$.
We find that the proposal quality is the best when $\tau_\text{conf}=0.5$.

\begin{table}[t]\centering
\setlength{\tabcolsep}{1.6mm}{
    \begin{tabular}{lcccc}  
    \hline
     & Vehicle & Bicycle & Chair & Motorbike\\     
    \hline \hline
    IoU ($\tau_\text{conf}\geqslant0.3$)& 57.8 & 84.3 & 93.0 & 83.3\\
    IoU ($\tau_\text{conf}\geqslant0.5$)& \textbf{70.7} & \textbf{90.6} & \textbf{98.0} & \textbf{86.9}\\
    IoU ($\tau_\text{conf}\geqslant0.7$)& 64.3 & 84.0 & \textbf{98.0} & 85.6\\
    \hline
  \end{tabular}}
  \caption{Average IoU of motion part proposals over different values of $\tau_\text{conf}$ on the Shape2Motion benchmark.}
\label{tab:Performance_all}\vspace{-8pt}
\end{table}

\paragraph{Effect of Motion Optimization Network (MON).}
To evaluate the effectiveness of MON,
we experiment an ablated version of our network without MON, over the two datasets.
The results are reported in row 3 of Table~\ref{tab:shape2motion} and \ref{tab:hudataset}.
By incorporating MON, our method achieves higher performance for
both motion part segmentation and motion attributes estimation, verifying its optimization effect.

\paragraph{Effect of Motion Part Proposal Module.}
In Figure~\ref{fig:Recall_over_threshold} (right), we evaluate the quality of motion part proposals via plotting the recall rate over average IoU. It can be observed that our method can generate many high-quality motion part proposals.
It can be seen that our method achieves very high recall rate ($>0.9$) for an IoU of $0.5$.

\begin{figure}[t]
  \centering
  \includegraphics[width=0.9\linewidth]{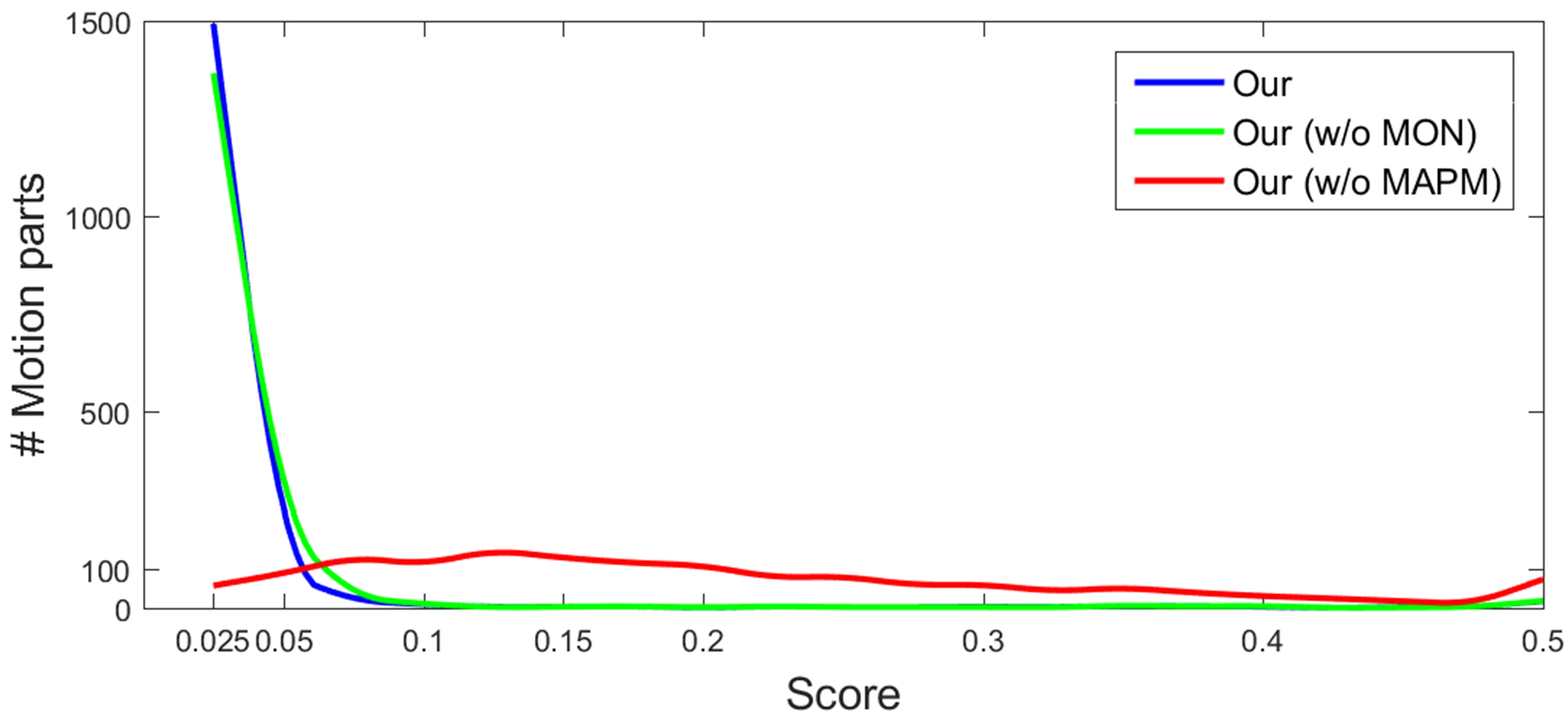}
  \caption{The distribution of motion part numbers over varying similarity scores (quality) of proposals.}
  \label{fig:match_error}\vspace{-5pt}
\end{figure}

\paragraph{Effect of Motion Attribute Proposal Module.}
To verify the effectiveness of this module, we compared to an ablated version of our method which replaces
this module by direct motion attributes regression as in the SGPN+BL in Table~\ref{tab:shape2motion}.
The quality of a predicted mobility is measured as the similarity between the mobility and its ground-truth, using similarity score in Equation (\ref{eq:match_score}).
In Figure~\ref{fig:match_error}, we plot the distribution of motion part numbers over varying similarity scores (quality).
The plots show that our full method works the best: The similarity of approximately $80\%$ mobility proposals is less than $0.02$.


\paragraph{Timing}
\label{sec:sect6_6}
The training of MPN and MON takes $31$ and $35$ hours for $50$ epoch on a NVIDIA TITIAN X GPU, respectively.
The testing time per 3D shape is $0.4$ seconds for MPN and $1$ seconds for MON.
The total computational cost is approximately $10$ seconds for each shape.

\section{Conclusion}

We have presented, Shape2Motion, an approach to simultaneous motion part segmentation and motion attribute estimation, using a single 3D shape as input.
The method adopts a proposal-and-optimize strategy, and consists of two deep neural networks,
i.e., a mobility proposal network (MPN) followed by a mobility optimization network (MON).
A key insight in training these networks is to fully exploit the coupling between a motion part and its mobility function, leading to a novel concept of motion-driven training, which may be valuable also for other scenarios.

\vspace{-5pt}
\paragraph{Limitations and future works.}
Our approach has a few limitations, which point out the directions of future study.
\kx{Representative failure cases can be found in the supplemental material.}
First, our method works with point cloud representation, which may not be able to represent shapes with highly detailed structure, such as a knife with scabbard. However, we believe the general framework of Shape2Motion can adapt to other shape representations.
%
Second, as a proposal-and-optimize framework, our method as a whole is not end-to-end trainable.
Third, our method does not support hierarchical mobility extraction, which is ubiquitous in real-world objects.
This involves higher-order analysis which is an interesting direction for future work.
We would also like to study the mobility analysis of scanned real-world objects.

\section*{Acknowledgement}
We thank the anonymous reviewers for their valuable comments. This work was supported in part by Natural Science Foundation of China (61572507, 61532003, 61622212),
and Natural Science Foundation of Hunan Province for Distinguished Young Scientists (2017JJ1002).

{\small
\bibliographystyle{ieee}
\bibliography{shape2motion}
}

\end{document}